\documentclass{article}
\usepackage{ijcai19}

\usepackage{times}
\usepackage{soul}
\usepackage{url}
\usepackage[hidelinks]{hyperref}
\usepackage[utf8]{inputenc}
\usepackage[small]{caption}
\usepackage{graphicx}
\usepackage{amsmath}
\usepackage{booktabs}
\usepackage{graphicx} 
\usepackage{color}
\usepackage{booktabs}
\usepackage{float}
\usepackage{url}
\usepackage{setspace} 
\usepackage[compact]{titlesec}
\usepackage{hyperref}

\graphicspath{ {./images/} }

\title{Predicting Confusion from Eye-Tracking Data with Recurrent Neural Networks\footnote{This work was presented at the 2nd Workshop on Humanizing AI (HAI) at IJCAI'19 in Macau, China.}}

\author{
Shane D. Sims\footnote{Contact Author}\and
Vanessa Putnam\and
Cristina Conati\\
\affiliations
The University of British Columbia\\
\emails
\{ssims, vputnam, conati\}@cs.ubc.ca
}

\begin{document}

\maketitle

\begin{abstract}
Encouraged by the success of deep learning in a variety of domains, we investigate the suitability and effectiveness of Recurrent Neural Networks (RNNs) in a domain where deep learning has not yet been used; namely detecting confusion from eye-tracking data.  Through experiments with a dataset of user interactions with ValueChart (an interactive visualization tool), we found that RNNs learn a feature representation from the raw data that allows for a more powerful classifier than previous methods that use engineered features. This is evidenced by the stronger performance of the RNN (0.74/0.71 sensitivity/specificity), as compared to a Random Forest classifier (0.51/0.70 sensitivity/specificity), when both are trained on an un-augmented dataset. However, using engineered features allows for simple data augmentation methods to be used. These same methods are not as effective at augmentation for the feature representation learned from the raw data, likely due to an inability to match the temporal dynamics of the data.
\end{abstract}

\section{Introduction}
An open problem in Affective Computing is that of detecting and adapting to a user's affective states in real-time. One such state is that of \textit{confusion}. When a user is confused while interacting with a system, they can experience a decrease in satisfaction, as well as performance \cite{yi2008visualized}. When this confusion is not resolved, the user may be unable to complete their task unless they are given some form of help. An obvious prerequisite to resolving a user's confusion is knowing that they are confused. A system that can detect when its user is confused (i.e. classify when the user is or is not confused) gains a human-like awareness that can be leveraged to provide appropriate interventions to resolve such confusion.    

Recurrent Neural Networks (RNNs) are deep learning (DL) methods that have achieved excellent results in domains with large amounts of available data \cite{karpathy2015deep}. Aside from detecting sentiment from text, work has been done using RNNs to recognize emotion from videos \cite{EbrahimiKahou:2015:RNN:2818346.2830596} and interaction events \cite{botelho2017improving}. The limited amount of work in this field (compared to the vast amount of work on object detection in pictures) is because data capturing affect is difficult to acquire and label. For example, one of the larger public emotion video datasets contains less than 1000 training items, with automated ground truth label generation \cite{dhall2017individual}. The data scarcity problem is exacerbated when relying on data from eye-trackers because it currently requires comparatively more specialized equipment and data collection in a laboratory setting. However, gaze has been shown to be a strong predictor of affective states such as boredom \cite{jaques2014predicting} and affective valence \cite{lalle2018prediction}. This fact, along with the success of deep learning in the other data scarce domains just mentioned is encouraging.

Prior work has shown that user confusion can be detected in the specific context of processing information visualizations using eye-tracking data such as gaze point, pupil changes, and the distance of the user's head from the screen. \cite{lalle2016prediction} used a Random Forest (RF) classifier on such data, achieving  \textit{confused} and \textit{not confused} class accuracies (i.e. sensitivity and specificity) of 57\% and 91\%, respectively. Their approach used engineered features based on domain knowledge about what is important for detecting confusion. We set out to determine if these results could be improved by using an RNN. This method could take advantage of the sequential nature of eye-tracking data (unlike previous approaches), while also learning a data-driven feature representation; a key difference between deep learning and other machine learning paradigms \cite{bengio2013representation}.

As the core contribution of this work, we show that RNNs learn a feature representation from raw eye-tracking data that allows for a more powerful classifier for detecting confusion than previous methods that used a Random Forest with engineered features (Section 6). This claim is supported by the results of our experiments, which show that RNN variants perform better than the RF before the dataset is augmented with synthetic examples. However, while simple data augmentation techniques have been applied to produce state of the art results with engineered features \cite{lalle2016prediction}, we discovered that these same techniques are less effective when applied to the raw eye-tracking dataset used to train the RNNs. Our results serve as a baseline upon which to evaluate our ongoing work in augmenting raw eye-tracking data. To the best of our knowledge, we are the first to use an RNN for the task of classifying a user state from raw eye-tracking data. A secondary contribution of this work is an account of our procedure (Section 5) for using RNNs to classify raw eye-tracking data from a small dataset; establishing some best practices along the way. In particular, we provide a simple method of increasing intra-class variance by cyclically splitting our data items into multiple items while maintaining the sequential structure. Our methods will be useful as more data of this type becomes available and research into related augmentation techniques progresses.

\section{Related Work}
Most work on predicting affect with deep learning has been in computer vision and natural language, where established methods already exist for classifying emotions from pictures, video, and sound \cite{EbrahimiKahou:2015:RNN:2818346.2830596} \cite{Xu:2016:VER:2911996.2912006} \cite{amer2014emotion}. These works focus on predicting one of anger, disgust, fear, happy, sad, surprise, or neutral.  There is less work that uses DL for emotion recognition based on data from live interaction with users,  which is ultimately what we want from an affect-sensitive artificial agent. 

Work that uses deep learning for classifying confusion has applied RNNs to sequential interaction data \cite{jiang2018expert} \cite{botelho2017improving}. These works seek to predict multiple emotions (one of which is confusion) while students interact with an Intelligent Tutoring System. While these works utilize interaction data, we utilize eye-tracking data, which has been shown to be a good predictor of emotional or attentional states such as mind wandering \cite{bixler2015automatic}, as well as boredom, and curiosity, while learning with educational software \cite{jaques2014predicting}. \cite{lalle2016prediction} predicted confusion using eye-tracking data, and achieved state of the art results using the Random Forest (RF) algorithm (Section 4). In \cite{pusiol2016vision}, a patient's gaze point was superimposed onto the face of a doctor from the patient's point of view. The result is a video that shows the patient's gaze point over time. An RNN was then used to predict a developmental disorder. To the best of our knowledge, deep learning has yet to be used for any affect predictive task based on eye-tracking data.

\section{Dataset}

The dataset used in this paper was collected from a user study using ValueChart, an interactive visualization for preferential choice \cite{lalle2016prediction}. Figure 1 shows an example of ValueChart configured for selecting rental properties from ten available alternatives (listed in the leftmost column), based on a set of relevant attributes (e.g., location or appliances). Although ValueChart has been evaluated for usability, the complexity of the decision tasks it supports means that it can still generate confusion \cite{yi2008visualized}\cite{conati2013adapt}. 

 \begin{figure}
\begin{center}
    \includegraphics[width=0.9\linewidth]{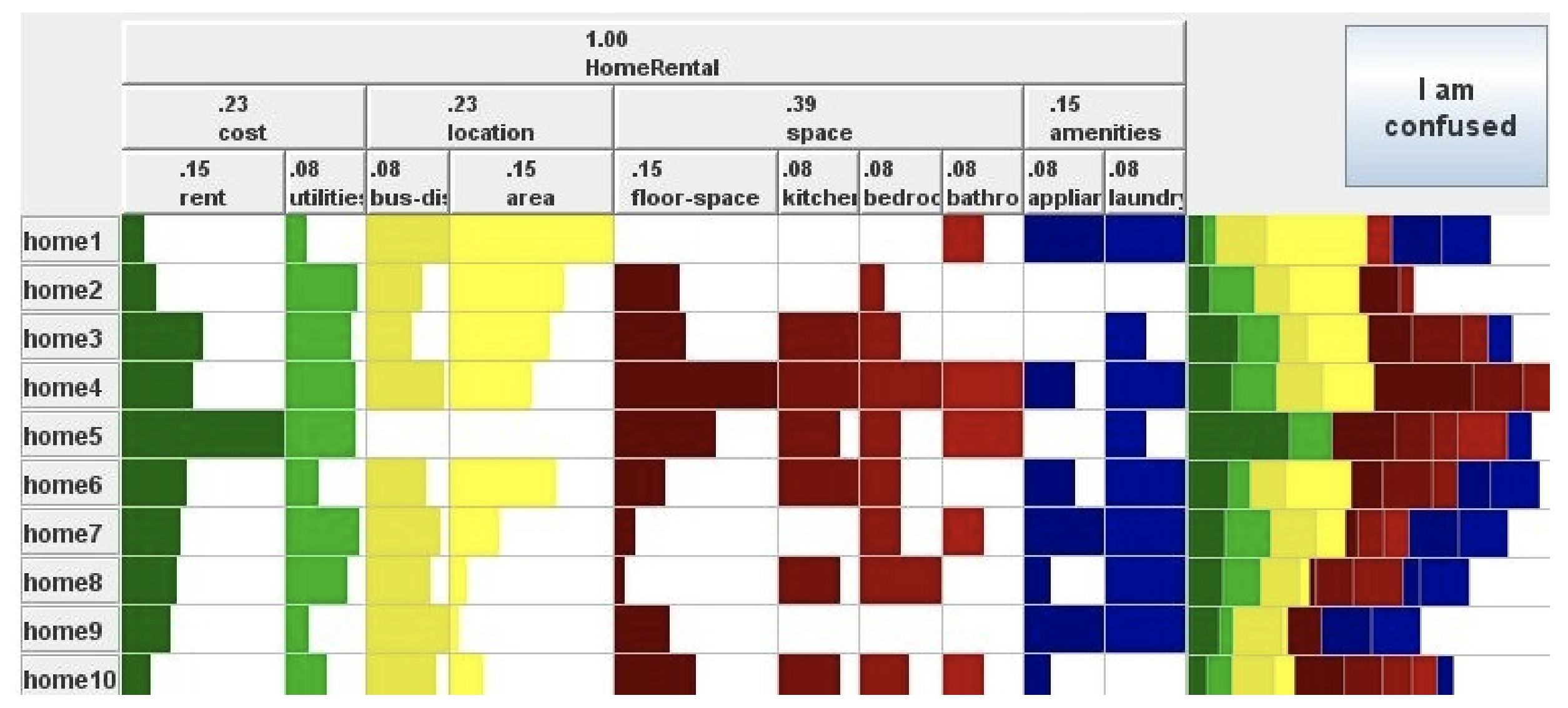}
\end{center}
   \caption{An example of the main elements of ValueChart.}
\end{figure}

In the user study where the data was collected, 136 participants were recruited to perform tasks with ValueChart. Each user performed a set of tasks relevant to exploring the available  options  for a given decision problem (e.g. selecting the best home based on the aggregation of price and size). There were 5 tasks, each of which was repeated 8 times (randomly ordered), resulting in 5440 task trials (136 users $\times5\times8$ trials). The user’s eyes were tracked with a Tobii T120 eye-tracker embedded in the study computer monitor. The study was administered in a windowless room with uniform lighting. Tasks have a mean duration of 13.7s (SD=11.3). To collect ground truth labels, users self-report their confusion by clicking on a button labelled “I am confused” (top right in Figure 1). To verify the labels, each participant was later shown replays of interaction segments centred around their confusion reports. Confusion was reported in 112 trials (2\%).

The eye-tracking data used for this prediction task can be expressed at three different levels of abstraction:

1. \textit{Raw eye-tracker sample}: data captured by the eye-tracker each time a sample is taken. At 120 Hz (the T120 sampling rate), a sample is collected 120 times per second. Each sample includes the (x,y) coordinates of the user’s gaze on the study screen, their pupil size, and the distance of each eye from the screen (Table 1). This data is rarely used in applications of eye-tracking, which typically rely on fixations.

2.  \textit{Fixation-based data}: eye-tracker proprietary software takes the raw eye-tracker data and generates sequences of fixations (gaze maintained at one point on the screen), saccades (quick movement of gaze from one fixation to another), the user's pupil size, and head distance from the screen. Tobii’s IV-T algorithm calculates this data [Olsen, 2012] by translating raw eye-tracker sample points to fixation locations by aggregating smaller eye movements that occur during fixations, such as tremors, drifts, and flicks.

3.  \textit{High-level features}: this level includes statistics of the user’s gaze (G), pupils (P), and head distance (HD). The G features are fixation rate and duration, saccade length, and relative and absolute saccade angles. Additional features relate to various Areas of Interest (AOI), such as fixation rate, longest duration, the proportion of fixations, and the number of transitions from a given AOI to every other. P and HD features are mean, SD, min, and max. The Eye Movement Data Analysis Toolkit (EMDAT) is a package capable of calculating such summative gaze statistics \cite{emdat}.
\begin{table}
\small
\centering
\begin{tabular}{l p{5cm}}
\toprule
\textbf{Feature} & \textbf{Definition}\\ 
\midrule
\textit{GazePointXLeft} & Horiz. screen position of gaze\\ 
\textit{GazePointYLeft} & Vert. screen position of gaze\\ 
\textit{CamXLeft} & Horiz. location of left pupil in the cam.\\
\textit{CamYLeft} & Vert. location of the left pupil in cam.\\
\textit{PupilLeft} & Size of the left pupil (mm)\\ 
\textit{DistanceLeft}& Dist. from eye-tracker to eye (mm)\\  
\textit{ValidityLeft} & Confidence that left eye identified\\
\bottomrule
\end{tabular}
\caption{ Raw eye-tracker features for the left eye.}
\end{table}

\section{Classifying Confusion with Random Forest from High-level Features} 

The state of the art approach to classifying confusion with the dataset just described uses a Random Forest classifier \cite{lalle2016prediction}. The model relies on the high-level features described in the previous section. Each trial in the dataset is represented as a 161-element vector encoding 149 gaze features, 6 pupil features, and 6 distance features, which summarize the user’s viewing and attentional behaviour for that trial.\footnote{The last second in each trial before a confusion self-report (or a pivot point) is removed to exclude signs of the intention to push the “I am confused” button from the data.} 

To address the highly imbalanced dataset, the Synthetic Minority Over-sampling Technique (SMOTE) \cite{chawla2002smote} was used to augment the minority class with synthetic examples. Their best results were obtained with a 200\% augmentation rate of their minority class and randomly downsampling the majority class to balance the training set.

In order to determine how much data leading up to an episode of confusion is needed to predict it, feature sets were built using two different windows of data. In the case where no confusion was reported in a trial, a randomly selected pivot point is used to provide a start point to the trial. A \textit{short window} captures 5s immediately before the end of a trial, while a \textit{full window} captures the whole trial. Results from these two window sizes were not statistically different.

To evaluate this setup,  Lall\'e et al. measured \textit{sensitivity} (proportion of \textit{confused} trials correctly identified as such) and \textit{specificity} (proportion of \textit{not confused} trials correctly identified as such). These metrics are also known as \textit{true positive} and \textit{true negative} rates, respectively. Lall\'e et al. achieved  0.57 sensitivity and 0.91 specificity after augmenting the dataset with SMOTE by 200\%. These results were achieved with a 75\% data validity requirement (i.e., ensuring each item was $\geq75$\%  valid). In both cases, the reported metrics were obtained over the unbalanced test set. Without using SMOTE the RF obtained a sensitivity of 0.51 and a specificity of 0.70. \footnote{ Lall\'e et al. also experimented with mouse-based interaction events. We focus on their eye-tracking results in order to keep our problem unimodal for the time being.}

\section{Classifying Confusion with RNNs and Raw Eye-tracker Data}

\noindent \textbf{Recurrent Neural Networks.}
Our approach is first distinguished from the one described above by our use of RNNs. Like other neural network approaches, RNNs learn a feature representation from data. It is important to note the distinction that deep neural networks are not just another classification algorithm (even if they can be used as such). In fact, the classification task is performed only at the last layers. The layers before this should be considered a representation learner, in that they serve to disentangle and discover the distinguishing features of the data fed into the model \cite{bengio2013representation}. This is in contrast to non-deep learning approaches that don't include this representation learning step. Non-deep learning models are \textit{given} a set of carefully selected features, the composition of which is generally based on human expertise about what is important for a given classification task. These models do not have to learn which features to use, they only have to learn \textit{from} the ones they are given.

RNNs are especially suited for sequential data (like ours) and the RNN variations we use are also more likely than others to learn long-term dependencies in sequential data, as they address the vanishing/exploding gradient problem common in the basic RNN when learning from long sequences \cite{Goodfellow-et-al-2016}.
An RNN learns its parameters by repeated application of a forward function. It is mainly the definition of this forward function that differentiates the RNN variants. When the output of the RNN is fully connected to a vector containing a node for each class, RNNs can be used for classification just like any other neural network.

We consider two RNN variations, namely Long-Short Term Memory (LSTM) networks \cite{hochreiter1997long} and Gated Recurrent Units (GRU) for their ability to learn from long sequences like those in our data. LSTMs are a gated variation of the basic RNN that differ in their use of self-loops to facilitate the learning of long term dependencies while also ensuring long term gradient flow. A GRU \cite{cho2014learning} is a gated RNN which is essentially a simplified LSTM that reduces the number of gates (and thus learnable parameters) used \cite{Goodfellow-et-al-2016}.

Given the primary strength of deep learning approaches being their representation learning ability, we use the raw eye-tracker sample data (Section 3) as input to our RNNs. We use data at this level of abstraction because it is the lowest level available and thus doesn't exclude any pattern expressed in the data captured by the eye-tracker. Any discriminator that could be lost in going to a higher level of data abstraction is necessarily maintained at this level. Previous work (Section 4) shows that summary statistics around fixations, saccades, pupil size, and head distance are strong features for classifying confusion. However, there may be other indicators in the data that we are not aware of. Providing data at the lowest level gives the model an opportunity to discover these \cite{bengio2013representation}. Each of our data items is, therefore, a 2D array with the number of rows corresponding to the number of samples captured in each trial. Each row within our items contains 14 columns; one for each feature in Table 1. 

\noindent \textbf{Pre-processing.}
We process the data before training our models by removing the last second of data and filtering out items shorter than 2s or with less than 65\% valid rows. We define a row to be invalid if both {\tt ValidityLeft} and {\tt ValidityRight} (Table 1) indicate that the eye-tracker is not confident in the data it has captured. We minimize the number of invalid values by identifying rows where at least one of {\tt ValidityLeft} or {\tt ValidityRight} is \texttt{True}, and then replace the invalid features with the valid ones, as feature values related to the left and right eye are similar at a given point in time. The steps described up to this point mimic part of the process that EMDAT uses to prepare high-level features \cite{emdat}. Applying these steps we discard 26 \textit{confused} and 1328 \textit{not confused} trials (similar to the number of trials discarded by EMDAT). All invalid values are replaced with -1 (a value not occurring in our data otherwise). 

While there is no fixed sequence length on which RNNs must operate, they typically require sequences shorter than those in the ValueChart confusion dataset (mean = 1644) \cite{neil2016phased}. \cite{lalle2016prediction} observed no significant difference when using the full sequence length compared to using the last 5s of data, which suggests that we can reduce our sequences down to at least 5s. We ran the RNNs on window sizes of 3, 4, 5, and 6s, and found no notable difference in performance. We thus chose to use a 5s window to strengthen comparability between our models and the Random Forest.

The next step we take to reduce sequence length comes after looking at the values in the rows of the raw eye-tracker samples. We observed that because of the high sampling rate of the eye-tracker, these values change only a small amount from one row to the next (i.e. after 8ms). Thus we split each sequence into 4 items with the same label by performing a cyclic partition (e.g. as when dealing a deck of cards), which preserves the temporal structure of the time series data (Figure 2). Splitting each data item like this reduces our sequence lengths by a factor of 4; the same amount of wall-clock time is now represented in a quarter of the number of rows.

\noindent \textbf{Data Augmentation. }
 \begin{figure}
\begin{center}
    \includegraphics[width=1.0\linewidth]{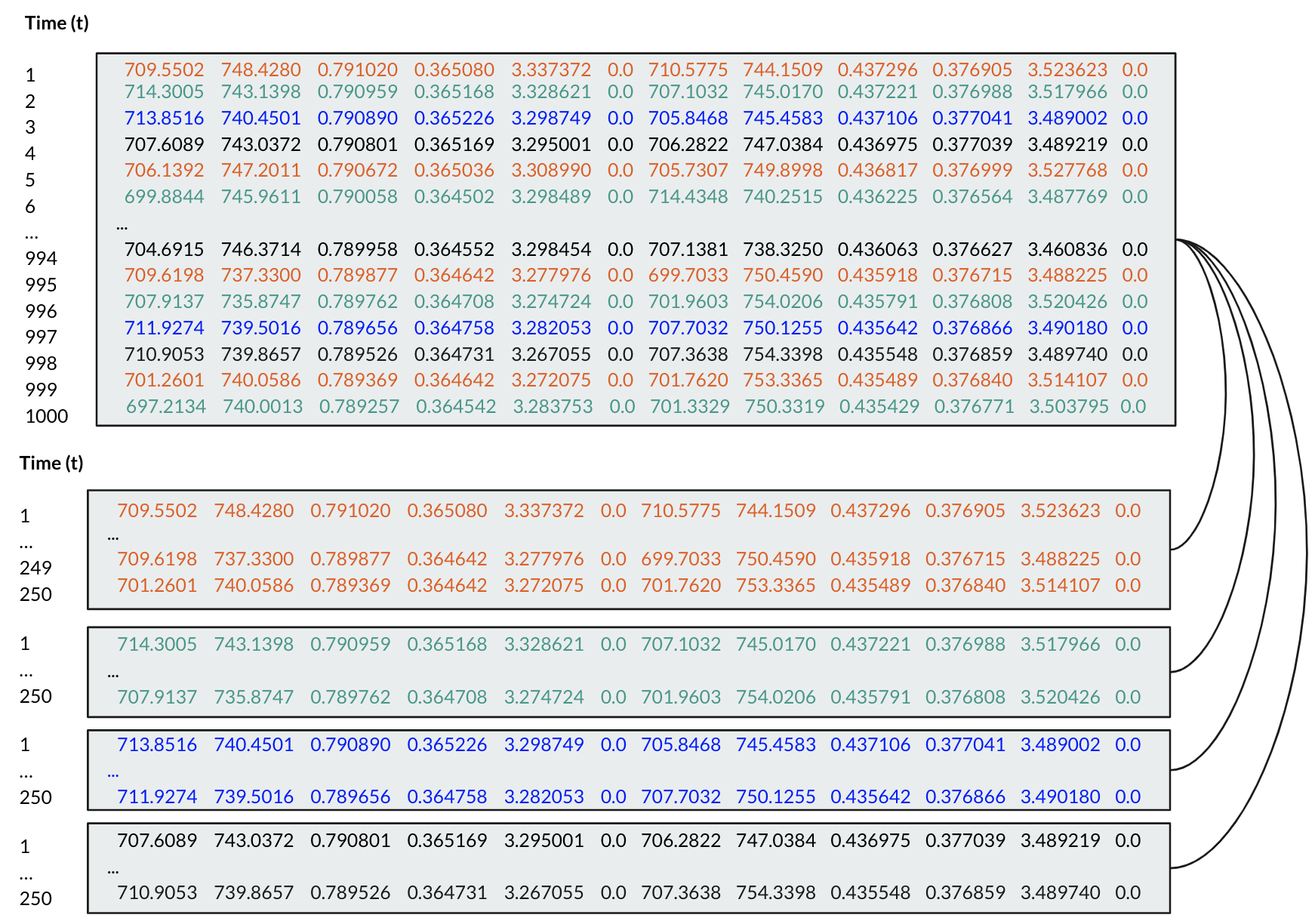}
\end{center}
   \caption{Cyclic partition of a single data item into four.} 
\end{figure}
The cyclic partition of the data serves a second function. By splitting the data we provide our model with multiple opportunities to learn from the same example in a more intelligent way than would be achieved by simply duplicating an example exactly (over-sampling). The difference between resulting items is enough to provide intra-class variance, while the cyclic partition ensures the preservation of the sequential pattern. It is thus our first attempt to address the small size of our dataset. However, the data-intensive nature of our approach and our small dataset necessitates augmentation beyond that achieved by the cyclic partition. Following the approach of previous work, we used SMOTE to increase the number of minority class items in our dataset.

\noindent \textbf{Implementation.} All of our models are implemented and trained using PyTorch. Limiting our models to a single layer and then following common heuristics for neural network design \cite{Goodfellow-et-al-2016}, we chose a hidden state of 256 units for our models. We use negative log likelihood as our loss function, with the Adam optimizer \cite{kingma2014adam} and an initial learning rate of $0.003$. We limit training to a maximum of 300 epochs, employing linear learning rate decay and early stopping to end training when validation performance stops improving. We use a batch size of $256$, on a single Nvidia GTX 1080 GPU. Metric calculations are done using the scikit-learn library \cite{scikit-learn}. To enhance reproducibility, our code is available at \url{https://github.com/sdv4/Predicting-Confusion-with-RNN_HAI-2019}.

\subsection{Experiment Setup and Evaluation.} 
To evaluate our models we employ across-user 10-fold Cross Validation (CV), which ensures that no user has contributed data items to both the training and test or validation sets of a given fold. We maintain the same ten data splits while training and evaluating each model and seed all random number generators in our code in order to enhance comparability.

In addition to sensitivity and specificity (defined in Section 4), we measure the \textit{Receiver Operating Characteristic (ROC) score}, which provides a unified measure of classifiers performance at a variety of decision thresholds. This allows us to choose a decision threshold that maximizes sensitivity and specificity. The threshold calculated is the one corresponding to the point closest to (0,1) on the ROC curve.  

Because of the highly imbalanced structure of our dataset (recall that validation and test sets are left unbalanced), sensitivity and specificity provide more meaningful metrics than accuracy. In the latter case, 98\% accuracy could be achieved by classifying everything as \textit{not confused}, since only 2\% of our data is labelled \textit{confused}.
Before training, we randomly down-sample \textit{not confused} items to balance the training set. 

For each CV fold, the dataset is split into training (90\%), validation (10\%) sets.
RNN model parameters (weights and biases) are updated via back-propagation on the loss over the training set. Validation set ROC is monitored to detect over-fitting and decide when to decrease the learning rate.

\section{Results}
Our most encouraging result is the performance of both the LSTM and the GRU against the Random Forest when no augmentation with synthetic data items is used. Here we see (in Table 2) that the RNNs exceed the Random Forest in terms of sensitivity while matching specificity. This result provides empirical evidence that the RNNs learn a more powerful feature representation with which to classify \textit{confused} items than does the Random Forest with high-level features. 

While both the GRU and LSTM achieve higher sensitivity scores than RF-SMOTE 200\%, the latter still outperforms the RNNs in terms of specificity. We postulate that this is because the Random Forest with 200\% confused class augmentation allows for triple the majority class items to be used for training the model, which allows for a more complete representation of the \textit{not confused} class in the training data. 

Our experiments using SMOTE (with 200\% augmentation) to increase the number of minority class training items for the LSTM and GRU showed no improvement in model performance. We also ran our experiments with the basic RNN, but as results were in all cases lower than those of the LSTM and GRU, we omit them from Table 2 for simplicity.

\begin{table}
\centering
\begin{tabular}{lrr} 
\toprule
\textbf{Classifier} &  \textbf{Sensitivity} & \textbf{Specificity}\\
\midrule
RF & 0.51 & 0.70 \\
RF-SMOTE 200\% & 0.57 & \textbf{0.91} \\
GRU & 0.70 & 0.70 \\
LSTM & \textbf{0.74} & 0.71\\
\bottomrule
\end{tabular}
\caption{Random Forest (RF) performance results of \protect\cite{lalle2016prediction}, compared to RNN validation set performance.}
\end{table}

Note that the results of Table 2 are calculated on the validation set. This is not an entirely unseen hold-out set, as it is monitored to detect over-fitting and ultimately to decide when to stop training. Validation set performance is often not a useful result to present, as it is easy to over-fit the model to data that (even indirectly) influences training. However, when we split the dataset into train (60\%), validation (30\%), and test (10\%) partitions, we found (Table 3) that the difference between the ROC score when calculated on the test set is close to that calculated on the validation set (average absolute value of difference is 0.0055). This suggests that our training procedure does not result in over-fitting to the validation set. We believe that in this case, reporting validation set performance is not only reasonable but also more useful in estimating the ability of the RNNs to discriminate between \textit{confused} and \textit{not confused} data since doing so allows for training with an extra 30\% of the data.

\begin{table}
\centering
\begin{tabular}{lrr} 
\toprule
\textbf{Classifier} & \textbf{Test ROC} & \textbf{Validation ROC}\\
\hline
GRU & 0.655 & 0.661 \\
LSTM & 0.653 & 0.658 \\
\bottomrule
\end{tabular}
\caption{Comparison of RNN performance on the validation set and the unseen test set.}
\end{table}

\section{Conclusion and Future Work}
In this paper, we have shown that RNNs can be used to classify confusion from raw eye-tracking data (i.e. without any manual feature engineering) and that the strength of this approach in capturing long term dependencies allows our results to approach the state of the art, even without effective data augmentation. When neither method is augmented with synthetic items, the RNNs outperform the Random Forest. When augmented, the Random Forest achieves a higher specificity but is still outperformed by the RNNs in terms of sensitivity. This provides evidence that the RNNs learn a more powerful feature representation from the raw data than is available in the high-level features used by the previous approach. These results also establish a baseline from which to measure future work in augmenting raw eye-tracking data. 

We have investigated initial data augmentation techniques, but have discovered that basic approaches to augmentation such as SMOTE do not work as well as they did for previous (non-sequential) approaches to classifying confusion. This is likely because while algorithms in the SMOTE family are necessarily distance based, they compute distance without matching the temporal dynamics of sequences \cite{gong2016model}. We have, however, identified promising lines of work for the task of augmenting our time series data, and have taken steps to develop and evaluate these. 

There are several limitations to be aware of when considering this work. First, while we support all of our claims with Cross Validated experimental results, we did not perform testing to verify the statistical significance of close results. Thus while it appears likely that the LSTM performed better than the GRU  (Table 2), we cannot be sure that this difference is statistically significant.

Throughout our work, we made a number of choices guided by intuition. Examples of this are the length of sequences used as input for our RNNs and the number of partitions in which to split our data. These choices would be better supported by significance testing in the first case and an analysis of variance in the latter. While we investigated window sizes near what previous work established as reasonable, it may be beneficial to use the entire sequence (the \textit{full window}), in the event that there are signals of confusion earlier on in a trial. In this current work, our choice of RNN variations limited our sequence size, but future work should include investigating the Phased LSTM, which is designed specifically for modelling long sequences \cite{neil2016phased}.

We performed almost no hyperparameter tuning on our RNNs or the learning algorithm. This is often a key element of success for deep learning models. A consequence of this is that our results may not reflect the true power of models, but rather provide a lower bound on what can be achieved. Future work in this respect will include optimizing depth, hidden unit size, adding dropout for regularization, and learning rate scheduling.

Another area to explore is using fixation-based features (the middle level of abstraction described in Section 3) with the same methods discussed in this paper. This may simplify data augmentation, due to the less complex structure of this data. The most critical area of future work is that of investigating and developing methods of data augmentation for raw eye-tracking data. We have implemented and experimented with affine transformations, with encouraging preliminary results. Following that, the effectiveness of Recurrent Variational Auto-Encoders \cite{chung2015recurrent} and Generative Adversarial Networks should be investigated for producing synthetic raw eye-tracking samples \cite{mogren2016c}.

While we are the first to classify a user affective state using deep learning methods on raw eye-tracking data, the work on classifying confusion has a longer history. Solving this task is critical if we are to build AI systems capable of expressing and responding to human emotions and mental states.
\subsection*{Acknowledgements}
We thank Sebastian Lall\'e for his patience and help in explaining his approach to predicting confusion from eye-tracker data. This work is supported in part by the Institute for Computing, Information and Cognitive Systems (ICICS) at UBC.
\newpage
\begin{spacing}{0.844}
\bibliographystyle{named} 
\bibliography{ijcai19}

\begin{thebibliography}{}

\bibitem[\protect\citeauthoryear{Amer \bgroup \em et al.\egroup
  }{2014}]{amer2014emotion}
Mohamed~R Amer, Behjat Siddiquie, Colleen Richey, and Ajay Divakaran.
\newblock Emotion detection in speech using deep networks.
\newblock In {\em 2014 IEEE international conference on acoustics, speech and
  signal processing (ICASSP)}, pages 3724--3728. IEEE, 2014.

\bibitem[\protect\citeauthoryear{Bengio \bgroup \em et al.\egroup
  }{2013}]{bengio2013representation}
Yoshua Bengio, Aaron Courville, and Pascal Vincent.
\newblock Representation learning: A review and new perspectives.
\newblock {\em IEEE transactions on pattern analysis and machine intelligence},
  35(8):1798--1828, 2013.

\bibitem[\protect\citeauthoryear{Bixler and
  D’Mello}{2015}]{bixler2015automatic}
Robert Bixler and Sidney D’Mello.
\newblock Automatic gaze-based detection of mind wandering with metacognitive
  awareness.
\newblock In {\em International Conference on User Modeling, Adaptation, and
  Personalization}, pages 31--43. Springer, 2015.

\bibitem[\protect\citeauthoryear{Botelho \bgroup \em et al.\egroup
  }{2017}]{botelho2017improving}
Anthony~F Botelho, Ryan~S Baker, and Neil~T Heffernan.
\newblock Improving sensor-free affect detection using deep learning.
\newblock In {\em International Conference on Artificial Intelligence in
  Education}, pages 40--51. Springer, 2017.

\bibitem[\protect\citeauthoryear{Chawla \bgroup \em et al.\egroup
  }{2002}]{chawla2002smote}
Nitesh~V Chawla, Kevin~W Bowyer, Lawrence~O Hall, and W~Philip Kegelmeyer.
\newblock Smote: synthetic minority over-sampling technique.
\newblock {\em Journal of artificial intelligence research}, 16:321--357, 2002.

\bibitem[\protect\citeauthoryear{Cho \bgroup \em et al.\egroup
  }{2014}]{cho2014learning}
Kyunghyun Cho, Bart Van~Merri{\"e}nboer, Caglar Gulcehre, Dzmitry Bahdanau,
  Fethi Bougares, Holger Schwenk, and Yoshua Bengio.
\newblock Learning phrase representations using rnn encoder-decoder for
  statistical machine translation.
\newblock {\em arXiv preprint arXiv:1406.1078}, 2014.

\bibitem[\protect\citeauthoryear{Chung \bgroup \em et al.\egroup
  }{2015}]{chung2015recurrent}
Junyoung Chung, Kyle Kastner, Laurent Dinh, Kratarth Goel, Aaron~C Courville,
  and Yoshua Bengio.
\newblock A recurrent latent variable model for sequential data.
\newblock In {\em Advances in neural information processing systems}, pages
  2980--2988, 2015.

\bibitem[\protect\citeauthoryear{Conati \bgroup \em et al.\egroup
  }{2013}]{conati2013adapt}
Cristina Conati, Enamul Hoque, Dereck Toker, and Ben Steichen.
\newblock When to adapt: Detecting user's confusion during visualization
  processing.
\newblock In {\em UMAP Workshops}. Citeseer, 2013.

\bibitem[\protect\citeauthoryear{Dhall \bgroup \em et al.\egroup
  }{2017}]{dhall2017individual}
Abhinav Dhall, Roland Goecke, Shreya Ghosh, Jyoti Joshi, Jesse Hoey, and Tom
  Gedeon.
\newblock From individual to group-level emotion recognition: Emotiw 5.0.
\newblock In {\em Proceedings of the 19th ACM international conference on
  multimodal interaction}, pages 524--528. ACM, 2017.

\bibitem[\protect\citeauthoryear{Ebrahimi~Kahou \bgroup \em et al.\egroup
  }{2015}]{EbrahimiKahou:2015:RNN:2818346.2830596}
Samira Ebrahimi~Kahou, Vincent Michalski, Kishore Konda, Roland Memisevic, and
  Christopher Pal.
\newblock Recurrent neural networks for emotion recognition in video.
\newblock In {\em Proceedings of the 2015 ACM on International Conference on
  Multimodal Interaction}, ICMI '15, pages 467--474, New York, NY, USA, 2015.
  ACM.

\bibitem[\protect\citeauthoryear{Gong and Chen}{2016}]{gong2016model}
Zhichen Gong and Huanhuan Chen.
\newblock Model-based oversampling for imbalanced sequence classification.
\newblock In {\em Proceedings of the 25th ACM International on Conference on
  Information and Knowledge Management}, pages 1009--1018. ACM, 2016.

\bibitem[\protect\citeauthoryear{Goodfellow \bgroup \em et al.\egroup
  }{2016}]{Goodfellow-et-al-2016}
Ian Goodfellow, Yoshua Bengio, and Aaron Courville.
\newblock {\em Deep Learning}.
\newblock MIT Press, 2016.
\newblock \url{http://www.deeplearningbook.org}.

\bibitem[\protect\citeauthoryear{Hochreiter and
  Schmidhuber}{1997}]{hochreiter1997long}
Sepp Hochreiter and J{\"u}rgen Schmidhuber.
\newblock Long short-term memory.
\newblock {\em Neural computation}, 9(8):1735--1780, 1997.

\bibitem[\protect\citeauthoryear{Jaques \bgroup \em et al.\egroup
  }{2014}]{jaques2014predicting}
Natasha Jaques, Cristina Conati, Jason~M Harley, and Roger Azevedo.
\newblock Predicting affect from gaze data during interaction with an
  intelligent tutoring system.
\newblock In {\em International conference on intelligent tutoring systems},
  pages 29--38. Springer, 2014.

\bibitem[\protect\citeauthoryear{Jiang \bgroup \em et al.\egroup
  }{2018}]{jiang2018expert}
Yang Jiang, Nigel Bosch, Ryan~S Baker, Luc Paquette, Jaclyn Ocumpaugh, Juliana
  Ma Alexandra~L Andres, Allison~L Moore, and Gautam Biswas.
\newblock Expert feature-engineering vs. deep neural networks: Which is better
  for sensor-free affect detection?
\newblock In {\em International Conference on Artificial Intelligence in
  Education}, pages 198--211. Springer, 2018.

\bibitem[\protect\citeauthoryear{Kardan}{}]{emdat}
Samad Kardan.
\newblock Eye movement data analysis toolkit.
\newblock \url{https://www.cs.ubc.ca/~skardan/EMDAT/#_Toc334603944}.
\newblock Accessed: 2019-06-03.

\bibitem[\protect\citeauthoryear{Karpathy and Fei-Fei}{2015}]{karpathy2015deep}
Andrej Karpathy and Li~Fei-Fei.
\newblock Deep visual-semantic alignments for generating image descriptions.
\newblock In {\em Proceedings of the IEEE conference on computer vision and
  pattern recognition}, pages 3128--3137, 2015.

\bibitem[\protect\citeauthoryear{Kingma and Ba}{2014}]{kingma2014adam}
Diederik~P Kingma and Jimmy Ba.
\newblock Adam: A method for stochastic optimization.
\newblock {\em arXiv preprint arXiv:1412.6980}, 2014.

\bibitem[\protect\citeauthoryear{Lall{\'e} \bgroup \em et al.\egroup
  }{2016}]{lalle2016prediction}
S{\'e}bastien Lall{\'e}, Cristina Conati, and Giuseppe Carenini.
\newblock Prediction of individual learning curves across information
  visualizations.
\newblock {\em User Modeling and User-Adapted Interaction}, 26(4):307--345,
  2016.

\bibitem[\protect\citeauthoryear{Lall{\'e} \bgroup \em et al.\egroup
  }{2018}]{lalle2018prediction}
S{\'e}bastien Lall{\'e}, Cristina Conati, and Roger Azevedo.
\newblock Prediction of student achievement goals and emotion valence during
  interaction with pedagogical agents.
\newblock In {\em Proceedings of the 17th International Conference on
  Autonomous Agents and MultiAgent Systems}, pages 1222--1231. International
  Foundation for Autonomous Agents and Multiagent Systems, 2018.

\bibitem[\protect\citeauthoryear{Mogren}{2016}]{mogren2016c}
Olof Mogren.
\newblock C-rnn-gan: Continuous recurrent neural networks with adversarial
  training.
\newblock {\em arXiv preprint arXiv:1611.09904}, 2016.

\bibitem[\protect\citeauthoryear{Neil \bgroup \em et al.\egroup
  }{2016}]{neil2016phased}
Daniel Neil, Michael Pfeiffer, and Shih-Chii Liu.
\newblock Phased lstm: Accelerating recurrent network training for long or
  event-based sequences.
\newblock In {\em Advances in neural information processing systems}, pages
  3882--3890, 2016.

\bibitem[\protect\citeauthoryear{Pedregosa \bgroup \em et al.\egroup
  }{2011}]{scikit-learn}
F.~Pedregosa, G.~Varoquaux, A.~Gramfort, V.~Michel, B.~Thirion, O.~Grisel,
  M.~Blondel, P.~Prettenhofer, R.~Weiss, V.~Dubourg, J.~Vanderplas, A.~Passos,
  D.~Cournapeau, M.~Brucher, M.~Perrot, and E.~Duchesnay.
\newblock Scikit-learn: Machine learning in {P}ython.
\newblock {\em Journal of Machine Learning Research}, 12:2825--2830, 2011.

\bibitem[\protect\citeauthoryear{Pusiol \bgroup \em et al.\egroup
  }{2016}]{pusiol2016vision}
Guido Pusiol, Andre Esteva, Scott~S Hall, Michael Frank, Arnold Milstein, and
  Li~Fei-Fei.
\newblock Vision-based classification of developmental disorders using
  eye-movements.
\newblock In {\em International Conference on Medical Image Computing and
  Computer-Assisted Intervention}, pages 317--325. Springer, 2016.

\bibitem[\protect\citeauthoryear{Xu \bgroup \em et al.\egroup
  }{2016}]{Xu:2016:VER:2911996.2912006}
Baohan Xu, Yanwei Fu, Yu-Gang Jiang, Boyang Li, and Leonid Sigal.
\newblock Video emotion recognition with transferred deep feature encodings.
\newblock In {\em Proceedings of the 2016 ACM on International Conference on
  Multimedia Retrieval}, ICMR '16, pages 15--22, New York, NY, USA, 2016. ACM.

\bibitem[\protect\citeauthoryear{Yi}{2008}]{yi2008visualized}
Ji~Soo Yi.
\newblock {\em Visualized decision making: development and application of
  information visualization techniques to improve decision quality of nursing
  home choice}.
\newblock PhD thesis, Georgia Institute of Technology, 2008.

\end{thebibliography}
\end{spacing}

\end{document}